\documentclass[11pt]{article}
\usepackage{amsmath}
\usepackage{amsfonts}
\usepackage[pdftex]{graphicx}
\usepackage{float}
\usepackage{mathrsfs}
\usepackage[framemethod=TikZ]{mdframed}
\usepackage{xcolor}
\usepackage{listofitems}

\newtheorem{theorem}{Theorem}[section]

\newtheorem{example}[theorem]{Example}

\newcommand{\be}{\begin{equation}}
\newcommand{\ee}{\end{equation}}

\newcommand{\myx}{\boldsymbol{x}}

\newcommand{\myE}{\mathbb{E}}

\newcommand{\mytheta}{\boldsymbol{\theta}}

\newcommand{\mymletheta}{\hat{\mytheta}}

\newcommand{\myTheta}{\boldsymbol{\Theta}}

\newcommand{\var}{x}

\newcommand{\newvar}{z}
\newcommand{\newVar}{Z} 

\newcommand{\model}[1]{%
 \readlist*\myargs{#1}%
  g(\myargs[1] | \myargs[2])}

\DeclareMathOperator*{\argmax}{argmax}

\newcommand{\qed}{\nobreak \ifvmode \relax \else
      \ifdim\lastskip<1.5em \hskip-\lastskip
      \hskip1.5em plus0em minus0.5em \fi \nobreak
      \vrule height0.75em width0.5em depth0.25em\fi}

\newcounter{example}
\renewcommand{\theexample}{\thesection.\arabic{example}}

\mdfdefinestyle{bluebox}{%
    backgroundcolor=blue!8,
    linecolor=blue,
    outerlinewidth=2pt,
    leftline=false,rightline=false,
    skipabove=\baselineskip,
    skipbelow=\baselineskip,
    frametitle=\mbox{},
}
\newmdenv[%
    style=bluebox,
    settings={\global\refstepcounter{bluebox}},
    frametitlefont={\bfseries Example~\theexample\quad},
]{bluebox}
\newmdenv[%
    style=bluebox,
    frametitlefont={\bfseries Example~\quad},
]{bluebox*}

\newmdenv[%
    backgroundcolor=red!8,
    linecolor=red,
    outerlinewidth=1pt,
    roundcorner=5mm,
    skipabove=\baselineskip,
    skipbelow=\baselineskip,
]{redbox}

\usepackage[english]{babel}
\usepackage{blindtext}

\begin{document}

  \title{\bf Implementing the ICE Estimator in Multilayer Perceptron Classifiers}
  \author{ \\
    Tyler Ward\thanks{Tyler Ward is an Adjunct Professor at NYU Tandon. E-mail: tw623@nyu.edu}
    }
  \maketitle

\pagebreak

\begin{abstract}

This paper describes the techniques used to implement the ICE estimator for a multilayer perceptron model, and reviews the performance of the resulting models. The ICE estimator is implemented in the Apache Spark MultilayerPerceptronClassifier, and shown in cross-validation to outperform the stock MultilayerPerceptronClassifier that uses unadjusted MLE (cross-entropy) loss. The resulting models have identical runtime performance, and similar fitting performance to the stock MLP implementations. Additionally, this approach requires no hyper-parameters, and is therefore viable as a drop-in replacement for cross-entropy optimizing multilayer perceptron classifiers wherever overfitting may be a concern.

\end{abstract}

\section{ICE Estimation}

The ICE methodology is described in detail in \cite{ICE}. This section contains a very brief review to introduce some common notation needed for the topics of this paper. 

Consider the model $ g(\myx_{i} | \mytheta) $ that assigns a probability to the regressors $ \myx_{i} $, and is parameterized by the parameters $ \mytheta $. Suppose the actual probability of $ \myx_{i} $ is $ f(\myx_{i}) $. Define the estimated log likelihood as 
\be
\label{mle}
\ell(\mytheta) := \frac{1}{n} \sum_{i=1}^{n} \log g(\myx_{i} | \mytheta)
\ee
and the expected log likelihood as 
\be
\mathcal{L}(\mytheta) :=\myE_{\newVar}[\log \model{\newVar, \mytheta}]
\ee
where the expectation is taken over the distribution $ f(\var) $. The maximum likelihood estimate $ \mymletheta $ is defined as
\be
\mymletheta := \argmax_{\mytheta \in \myTheta}\ell(\mytheta).
\ee

We define the negative expected Hessian matrix 

\be
J(\mytheta) := - \myE_{\newVar}[\partial^2_{\mytheta}\log \model{\newVar, \mytheta}]=-\int f(\newvar) \partial^2_{\mytheta} \log \model{\newvar,\mytheta} d\newvar,
\ee

and the Fisher Information matrix 

\be
I(\mytheta):=\myE_{\newVar}[\partial_{\mytheta}\log \model{\newVar, \mytheta} \partial_{\mytheta^T}\log \model{\newVar, \mytheta}].
\ee

In \cite{ICE}, the ICE objective is defined as 

\be
-\ell^{*}(\mytheta) = -\ell(\mytheta) + \frac{1}{n} tr(\hat{I}\hat{J}^{-1})
\ee

Taking $ v_{(\mytheta, \var)} = - \partial_{\mytheta} \ell(\mytheta, \var) $, this can be rewritten as 

\be
\label{iceRaw}
-\ell^{*}(\mytheta) = -\ell(\mytheta) + \frac{1}{n} \sum_{i} v^{T}_{(\mytheta, \var_{i})} (\hat{J}^{-1}) v_{(\mytheta, \var_{i})}
\ee

As explained in \cite{ICE}, optimizing to this objective instead of Equation \ref{mle} would eliminate the leading order bias terms. This reduces the cross-validation or prediction bias from $ O(\frac{1}{N}) $ to $ O(\frac{1}{N^{3/2}}) $ as a function of the observation count $ N $. For realistic data set sizes, this should effectively eliminate overfitting without requiring cross validation or any hyper-parameters.

This computation requires $ O(d^{3}) $ time and $ O(d^{2}) $ space if $ d $ is the number of parameters in $ \mytheta $. This equation is therefore well defined, but is totally unsuitable for numerical computation, having two main problems: 

\begin{enumerate}
\item For even moderate parameter counts (e.g. 20+) the inverse condition number of $ \hat{J} $ is typically less than machine precision. Therefore, the direct numerical computation of this quantity will be quickly dominated by numerical errors, even when a stabilized SVD based pseudo-inversion is used. 
\item The computational cost of the inversion of $ \hat{J} $ is $ O(d^{3}) $, and this must be done on every iteration within the optimizer during model fitting. This quickly becomes intractable for parameter counts beyond a few hundred. This is less severe than the first issue, but still a major impediment to wide scale adoption in the highly parameterized models where bias is most an issue. 
\end{enumerate}

\section{Efficient Approximations}

Several efficient approximations of Equation \ref{iceRaw} have been proposed, here we consider the following:

\be
\label{iceApprox}
-\ell^{*}(\mytheta) = -\ell(\mytheta) + \frac{1}{n} \sum_{i} v^{T}_{(\mytheta, \var_{i})} (\hat{D}^{-1}) v_{(\mytheta, \var_{i})}
\ee

An approximation for the gradient Equation \ref{iceRaw} is 

\be
\label{iceGrad3}
-\partial_{\mytheta} \ell^{*}(\mytheta) \approx v_{(\mytheta, \var)} + \frac{2}{n} \left[ \sum_{i} v_{(\mytheta, \var_{i})} \left[v^{T}_{(\mytheta, \var_{i})} \hat{D}^{-1}  v_{(\mytheta, \var_{i})} \right] \right] 
\ee

\subsection{Numerical Stabilization}

Both Equation \ref{iceApprox} and Equation \ref{iceGrad3} can suffer from the potential singularity (or ill conditioning) of $ \hat{D} $. 

To improve numerical stability, we truncate to zero any gradient element with $ \| [v_{(\mytheta, \var_{i})}]_{k} \| < \sqrt{\varepsilon} * max_{k}( \| [v_{(\mytheta, \var_{i})}]_{k} \|)  $, where here $ \varepsilon $ is double precision machine epsilon, approximately $ 10^{-16} $. These terms are too small to change the outcome of a dot product within machine error. A vector so truncated is indistinguishable via dot products from one that has not been, however it is possible for such terms to add numerical instability due to rounding errors in the computation of $ \hat{D}^{-1} $. Similarly, for each element $ [\hat{D}]_{k} $ of $ \hat{D} $, the value of $ [\hat{D}^{-1}]_{k} $ is computed as

\be
\label{objectiveStabilizer}
 [\hat{D}^{-1}]_{k} \approx \frac{1}{w[\hat{D}]_{k} + (1 - w)[v_{(\mytheta, \var_{i})}]_{k}^{2} }
\ee 

Where the weight $ w $ is computed as 

\be
w_{k} = e^{-\frac{\sqrt{\varepsilon} max_{j}(\| [\hat{D}]_{j} \|)}{ max(0, [\hat{D}]_{k})}}
\ee

This weight is a continuous function of $ [\hat{D}]_{k} $, and goes to zero as $ [\hat{D}]_{k} $ becomes small enough that it is dominated by rounding errors. In addition, this will pin the term in question at $ 1.0 $ for negative values of $ [\hat{D}]_{k} $, thus preventing instability from forming when the optimizer is not near the MLE solution and $ \hat{D} $ is not positive definite. 

\section{Neural Network Implementation}

For implementation within neural networks, it is necessary to be able to compute $ \hat{D} $ using backpropagation. The techniques for performing this computation are described by Le Cun in \cite{LeCun89}, section 3.2. Another description of this approach is given by \cite{Buntine94}, Section 4.1. 

Consider the neural network with $ L $ layers, and cost function $ C $. Assume also that no connections skip layers. Typically, for a classifier, $ C $ would be a cross entropy loss, with $ y $ being the known labels of the training data. 

\be
C(y, f_{L}(W_{L}f_{L-1}(W_{L-1} ... f_{1}(W_{1}x))))))
\ee

Here $ W_{l} $ is the matrix of edge weights for layer $ l $, and it is assumed that the activation function at each layer is uniform, hence univariate function $ f_{l} $ when applied to a vector argument produces a vector output. We may define the activation of layer $ l $ as 

\be
a_{0} = x
\ee

\be
a_{l} = f_{l}(W_{l}f_{l-1}(W_{l-1}... f_{1}(W_{1}x)))) = f_{l}(W_{l}a_{l-1})
\ee

Then we can rewrite the neural network, ignoring the parameter $ y $, as

\be
C(a_{L})
\ee

Note that the weights contain an implicit bias term, so more explicitly, the activation of the $ i $'th node in layer $ l $ would be 

\be
(a_{l})_{i} = f_{l}((W_{l})_{(i, 0)} + \sum_{k} (a_{l-1})_{k}(W_{l})_{(i,k)})
\ee

Then the second derivative of the objective function $ C $ of the network may be constructed by inverting this sum (so it runs over $ i $ that is connected to by $ k $). 

\be
\label{eq:hessRaw}
\frac{\partial^{2}C}{\partial (a_{l-1})^{2}_{k}} = \sum_{i} \left[ \frac{\partial^{2}C}{\partial (a_{l})^{2}_{i}} ((f'_{l}(W_{l}a_{l-1}))_{i}(W_{l})_{(i,k)})^{2} + \frac{\partial C}{\partial (a_{l})_{i}} (f_{l}''(W_{l}a_{l-1}))_{i}(W_{l})^{2}_{(i, k)} \right]
\ee

Where here the derivatives $ f'_{l} $ and $ f''_{l} $ are taken with respect to the function's sole argument. Note that this equation is only accurate for the case where the off diagonal elements of $ \frac{\partial^{2}C}{\partial (a_{l-1})^{2}_{k}} $ are actually zero. If any are nonzero (as would be the case in practice), then this equation is only an approximation.  Renaming this quantity

\be
(u'_{l})_{i} = f'_{l}(W_{l}a_{l-1})
\ee

and 

\be
(u''_{l})_{i} = f''_{l}(W_{l}a_{l-1})
\ee

Equation \ref{eq:hessRaw} may be rewritten as

\be
\label{eq:hessRaw2}
\frac{\partial^{2}C}{\partial (a_{l-1})^{2}_{k}} = \sum_{i} \left[ \frac{\partial^{2}C}{\partial (a_{l})^{2}_{i}} (u'_{l})^{2}_{i} + \frac{\partial C}{\partial (a_{l})_{i}} (u_{l}'')_{i}  \right] (W_{l})^{2}_{(i, k)} 
\ee

The derivatives with respect to the weights are then. 

\be
\label{eq:hessFinal}
\frac{\partial^{2}C}{\partial (W_{l})^{2}_{(i, k)}} = \left[ \frac{\partial^{2}C}{\partial (a_{l})^{2}_{i}} (u'_{l})_{i}^{2} + \frac{\partial C}{\partial (a_{l})_{i}} (u''_{l})_{i} \right] (a_{l-1})^{2}_{k} 
\ee

These formulas are then suitable for a backpropagation implementation.

\subsection{Backpropagation Implementation}

For modern neural nets, derivatives must be computed using backpropagation for efficiency reasons. This section describes the backpropagation techniques used to compute first and second derivatives. 

\subsubsection{Backpropagation Gradient Implementation}

Considering the network as previously defined, we may define the auxillary value 

\be
\delta_{L} = (u'_{L}) \cdot \nabla_{a_{L}} C
\ee

and then recursively define it for all other layers. 

\be
\delta_{l-1} = (u'_{l-1})(W_{l})^{T} \delta_{l}
\ee

We then may compute the gradient of $ C $ using these values. 

\be
\label{eq:backprop}
\nabla_{W_{l}} C = (\delta_{l})(a_{l-1})^{T}
\ee

For future reference, note that 

\be
\label{partials}
\frac{\partial C}{\partial (a_{l-1})} = \nabla_{a_{l-1}} C = (\delta_{l}) (W_{l})^{T}
\ee

and that $ \frac{\partial C}{\partial (a_{L})} $ is directly computable from the definition of $ C $.

\subsubsection{Backpropagation Hessian Diagonal Implementation}

Analogously to the definition of $ \delta $, we define an auxillary value $ \gamma $ using Equation \ref{eq:hessRaw}. Because this will be computing only the diagonal of the hessian, it is necessary to write it summation form.  

\begin{eqnarray}
(\gamma_{l-1})_{k} &=& \frac{\partial^{2}C}{\partial (a_{l-1})^{2}_{k}}  \\
                               &=& \sum_{i} \left[ \frac{\partial^{2}C}{\partial (a_{l})^{2}_{i}} (u'_{l})_{i}^{2} + \frac{\partial C}{\partial (a_{l})_{i}} (u_{l}'')_{i}\right] (W_{l})^{2}_{(i, k)} \\
                               &=& \sum_{i} \left[ (\gamma_{l})_{i} (u'_{l})^{2}_{i} + \frac{\partial C}{\partial (a_{l})_{i}} (u_{l}'')_{i} \right] (W_{l})^{2}_{(i, k)}  \label{eq:gamma}
\end{eqnarray}

and that $ (\gamma_{L}) = \frac{\partial^{2}C}{\partial (a_{L})^{2}} $ is computable directly from the definition of $ C $.  Additionally, $ \frac{\partial C}{\partial (a_{l})_{i}}  $ may be computed using equation \ref{partials}. 

Then the diagonal of the hessian itself is computed using Equation \ref{eq:hessFinal}. 

\be
\label{eq:hessFinal2}
\frac{\partial^{2}C}{\partial (W_{l})^{2}_{(i, k)}} = \left[ (\gamma_{l})_{i} (u'_{l})_{i}^{2} + \frac{\partial C}{\partial (a_{l})_{i}} (u''_{l})_{i} \right] (a_{l-1})^{2}_{k}
\ee

The combination of equations \ref{eq:gamma}, \ref{eq:hessFinal2}, and \ref{eq:backprop} are sufficient to compute the first and (non-mixed) second derivatives of the neural network in a single backpropagation pass. 

Recall that Equation \ref{eq:hessRaw} is only an approximation. If more accuracy is needed (at the expense of more computation), then the matrix $ (\Gamma_{l})_{(i, k)} $ (instead of the vector $(\gamma_{l})_{i}$) may be computed and backpropagated using a similar formula. In which case Equation \ref{eq:hessFinal2} relies instead on $ (\Gamma_{l})_{(i, i)} $ but is otherwise unchanged. That analysis is beyond the scope of the current work. 

\subsection{Derivatives of Cross-Entropy Multinomial Logistic Loss}

Suppose the loss function  $ C $ is cross entropy loss using a multinomial logistic (i.e. softmax) classifier. Defining the vector valued multinomial logistic function as 

\be
(L(a_{L}))_{i} =  \frac{ exp((a_{L})_{i}) }{\sum_{k} exp((a_{L})_{k})}
\ee

Then the cross entropy loss of a single observation is

\be
C(y, (a_{L}))_{i} = - y_{i} \ln [\frac{ exp((a_{L})_{i}) }{\sum_{k} exp((a_{L})_{k})} ] = - y_{i} \ln [(L(a_{L}))_{i}]
\ee

Where $ y $ is a one-hot encoding of the classes for the given observation. The derivatives of this loss function with respect to $ a_{L} $ are

\be
\frac{\partial C(y, a_{L})}{\partial (a_{L})_{i}}  = (L(a_{L})_{i} - y_{i})
\ee

and 

\be
(\frac{\partial^{2}C}{\partial (a_{L})_{i}^{2}})_{i} =\frac{\partial }{\partial (a_{L})_{i}} [L(a_{L})_{i} - y_{i}] = [1 - L(a_{L})_{i}] L(a_{L})_{i}
\ee

Note that traditionally, a multilayer perceptron will use the identity activation function for the last layer, in which case $ f_{L}(x) = x $.

\section{Results}

The ICE estimator was implemented in the Apache Spark MultilayerPerceptronClassifier, and compared against a stock MultilayerPerceptronClassifier using Spark version 2.4.5. This implementation was chosen due to the dominant marketshare of Spark and the ease of implementation and testing within that codebase. Because the Spark MLP model does not provide for regularization or drop-out, this approach could not be compared against those approaches within this codebase. 

The computation was performed on a selection of Freddie Mac mortgage data made publicly available through the Single Family Loan Level Dataset. 

\subsection{Mortgage Data}

The mortgage data chosen is a sample of approximately 2 million loan-month observations from Freddie Mac fixed rate, mortgages originated in 2001. Loan months are selected such that each loan is current on all payments at the start of the month, and then has the potential to prepay the loan, remain current, or miss a payment. Therefore, a classifier is constructed over these three outcomes. From this data, 11 regressors are chosen.

\begin{itemize}
\item Loan age in months.
\item Mark-to-market loan-to-value ratio.
\item Loan prepayment incentive.
\item Borrower credit score.
\item Indicator for first-time-buyer.
\item Unit count.
\item Combined Loan-to-Value at origination.
\item Debt-to-Income at origination.
\item Unpaid balance.
\item Interest rate.
\item Indicator for prepayment penalties. 
\end{itemize}

The exact definition of these regressors is beyond the scope of this paper, but this represents a broad set of applicable regressors for a typical loan. It includes some highly unbalanced regressors (such as unit count), and also some highly co-linear regressors (such as Mark-to-Market and combined Loan-to-Value ratios). All regressors are demanded to be non-negative, except for loan-to-value regressors which are required to be strictly positive, and incentive which is required to be between -1.0 and 1.0 to remove a handful of loans with data entry errors. This filtering removes less than 0.5\% of the data. 

The data is randomly split between fitting and testing datasets using probabilities (0.25, 0.75). The fitting set is further reduced to the target size for each run, and the entire testing set of approximately 1.5MM observations is used for cross validation of the resulting models in each run. 

\subsection{Performance}

For this section, performance was tested on four layer configurations. Each model has 11 input regressors and 3 classification states. The models tested are described in Table \ref{tab:config}.

\begin{table}[!h]
\begin{center}
\begin{tabular}{|ccl|}
\hline
Layer Configuration & Parameter Count & Description \\
\hline
$ [11,3] $ & 36 & The simplest model, with no hidden layers \\
$ [11,5,3] $ & 78 & A model with a single hidden layer 5 neurons wide. \\
$ [11, 8, 5, 3] $ & 159 & A model with two hidden layers. \\
$ [11, 11, 8, 5, 3] $ & 291 & A model with three hidden layers. \\
\hline
\end{tabular}
\caption{The model configurations.}
\label{tab:config}
\end{center}
\end{table}

Each configuration was fit $ 10 $ times on randomly drawn fitting sets of various sizes using both MLE and ICE. The cross-entropy on the testing set was averaged for each series of tests. All optimization are performed using l-bfgs, which generally produced better fits in all the tests. The results are presented in Tables \ref{tab:small}, \ref{tab:medium}, \ref{tab:large}, and \ref{tab:xl}.

\begin{table}[!h]
\begin{center}
\begin{tabular}{|ccccc|}
\hline
Fitting Set Size & ICE (test) & ICE (fit) & MLE (test) & MLE (fit)  \\
\hline
 $ 128 $ & $ 0.954571 $ & $ 0.155540 $ & $ 4.419692 $ & $ 0.103623 $ \\
 $ 256 $ & $ 0.274862 $ & $ 0.229512 $ & $ 0.505011 $ & $ 0.190171 $ \\
 $ 512 $ & $ 0.244429 $ & $ 0.205278 $ & $ 0.278151 $ & $ 0.182650 $ \\
 $ 1024 $ & $ 0.228834 $ & $ 0.203445 $ & $ 0.232094 $ & $ 0.191234 $ \\
 $ 2048 $ & $ 0.222566 $ & $ 0.199967 $ & $ 0.219652 $ & $ 0.194121 $ \\
 $ 4096 $ & $ 0.213989 $ & $ 0.203505 $ & $ 0.211550 $ & $ 0.201573 $ \\
 $ 8192 $ & $ 0.205199 $ & $ 0.199506 $ & $ 0.204879 $ & $ 0.198632 $ \\
 $ 16384 $ & $ 0.204458 $ & $ 0.201699 $ & $ 0.204206 $ & $ 0.201342 $ \\
 $ 32768 $ & $ 0.202891 $ & $ 0.201684 $ & $ 0.202966 $ & $ 0.201445 $ \\
 $ 65536 $ & $ 0.201936 $ & $ 0.200623 $ & $ 0.201984 $ & $ 0.200534 $ \\
 $ 131072 $ & $ 0.201702 $ & $ 0.201166 $ & $ 0.201915 $ & $ 0.201027 $ \\
\hline
\end{tabular}
\caption{Cross entropy loss for configuration $ [11,3] $ ($36$ parameters) }
\label{tab:small}
\end{center}
\end{table}

\begin{table}[!h]
\begin{center}
\begin{tabular}{|ccccc|}
\hline
Fitting Set Size & ICE (test) & ICE (fit) & MLE (test) & MLE (fit)  \\
\hline
 $ 128 $ & $ 0.272454 $ & $ 0.229064 $ & $ 10.900275 $ & $ 0.023882 $ \\
 $ 256 $ & $ 0.265012 $ & $ 0.243661 $ & $ 10.647536 $ & $ 0.060359 $ \\
 $ 512 $ & $ 0.250343 $ & $ 0.230830 $ & $ 7.500989 $ & $ 0.093122 $ \\
 $ 1024 $ & $ 0.227124 $ & $ 0.220959 $ & $ 3.323920 $ & $ 0.132962 $ \\
 $ 2048 $ & $ 0.221264 $ & $ 0.227862 $ & $ 1.003692 $ & $ 0.167186 $ \\
 $ 4096 $ & $ 0.213282 $ & $ 0.213359 $ & $ 0.568460 $ & $ 0.178987 $ \\
 $ 8192 $ & $ 0.209258 $ & $ 0.205357 $ & $ 0.204754 $ & $ 0.182946 $ \\
 $ 16384 $ & $ 0.206359 $ & $ 0.206933 $ & $ 0.197334 $ & $ 0.191357 $ \\
 $ 32768 $ & $ 0.198800 $ & $ 0.197964 $ & $ 0.194360 $ & $ 0.191656 $ \\
 $ 65536 $ & $ 0.197006 $ & $ 0.195556 $ & $ 0.193498 $ & $ 0.191018 $ \\
 $ 131072 $ & $ 0.194683 $ & $ 0.195296 $ & $ 0.192713 $ & $ 0.193312 $ \\
\hline
\end{tabular}
\caption{Cross entropy loss for configuration $ [11,5,3] $ ($78$ parameters) }
\label{tab:medium}
\end{center}
\end{table}

\begin{table}[!h]
\begin{center}
\begin{tabular}{|ccccc|}
\hline
Fitting Set Size & ICE (test) & ICE (fit) & MLE (test) & MLE (fit)  \\
\hline
 $ 128 $ & $ 0.318874 $ & $ 0.321617 $ & $ 10.857075 $ & $ 0.024558 $ \\
 $ 256 $ & $ 0.291447 $ & $ 0.268866 $ & $ 9.277588 $ & $ 0.030094 $ \\
 $ 512 $ & $ 0.260094 $ & $ 0.261958 $ & $ 4.986087 $ & $ 0.044398 $ \\
 $ 1024 $ & $ 0.237706 $ & $ 0.239615 $ & $ 2.453838 $ & $ 0.090196 $ \\
 $ 2048 $ & $ 0.225294 $ & $ 0.227843 $ & $ 1.055846 $ & $ 0.115964 $ \\
 $ 4096 $ & $ 0.217699 $ & $ 0.220031 $ & $ 0.636030 $ & $ 0.156975 $ \\
 $ 8192 $ & $ 0.213616 $ & $ 0.210857 $ & $ 0.332622 $ & $ 0.169217 $ \\
 $ 16384 $ & $ 0.211100 $ & $ 0.210478 $ & $ 0.212039 $ & $ 0.182773 $ \\
 $ 32768 $ & $ 0.209711 $ & $ 0.209761 $ & $ 0.196397 $ & $ 0.189368 $ \\
 $ 65536 $ & $ 0.201107 $ & $ 0.199574 $ & $ 0.193728 $ & $ 0.189233 $ \\
 $ 131072 $ & $ 0.199561 $ & $ 0.199458 $ & $ 0.192666 $ & $ 0.191408 $ \\
\hline
\end{tabular}
\caption{Cross entropy loss for configuration $ [11,8,5,3] $ ($159$ parameters) }
\label{tab:large}
\end{center}
\end{table}

\begin{table}[!h]
\begin{center}
\begin{tabular}{|ccccc|}
\hline
Fitting Set Size & ICE (test) & ICE (fit) & MLE (test) & MLE (fit)  \\
\hline
 $ 128 $ & $ 0.305902 $ & $ 0.296990 $ & $ 7.391331 $ & $ 0.011194 $ \\
 $ 256 $ & $ 0.319394 $ & $ 0.328911 $ & $ 7.926534 $ & $ 0.023455 $ \\
 $ 512 $ & $ 0.260440 $ & $ 0.256655 $ & $ 2.933313 $ & $ 0.034358 $ \\
 $ 1024 $ & $ 0.236597 $ & $ 0.242894 $ & $ 2.112070 $ & $ 0.045172 $ \\
 $ 2048 $ & $ 0.225582 $ & $ 0.223973 $ & $ 1.262838 $ & $ 0.083218 $ \\
 $ 4096 $ & $ 0.218349 $ & $ 0.216054 $ & $ 0.720304 $ & $ 0.116865 $ \\
 $ 8192 $ & $ 0.216059 $ & $ 0.217507 $ & $ 0.367875 $ & $ 0.153935 $ \\
 $ 16384 $ & $ 0.212912 $ & $ 0.210730 $ & $ 0.234384 $ & $ 0.172002 $ \\
 $ 32768 $ & $ 0.211188 $ & $ 0.206361 $ & $ 0.202727 $ & $ 0.180060 $ \\
 $ 65536 $ & $ 0.210527 $ & $ 0.209783 $ & $ 0.195163 $ & $ 0.188557 $ \\
 $ 131072 $ & $ 0.206824 $ & $ 0.205842 $ & $ 0.192954 $ & $ 0.189868 $ \\
\hline
\end{tabular}
\caption{Cross entropy loss for configuration $ [11,11,8,5,3] $ ($291$ parameters) }
\label{tab:xl}
\end{center}
\end{table}

In all four configuratons, ICE effectively eliminates overfitting, whereas MLE suffers severe overfitting for smaller sample sizes. In all four tests MLE performs slightly better with very large sample sizes, but the difference is not very large. For $ [11,3] $ configuration shown in Table \ref{tab:small}, ICE shows some overfitting, but much less than MLE, whereas for the other configurations no material amount of overfitting is present. This is likely due to the specifics of the l-bfgs fitting algorithm, which can generally search the parameter space much more efficiently for a more nearly linear model such as $ [11,3] $ than it can for more complicated configurations. The bias reduction from ICE is asymptotic, so it is not surprising that the approach is weaker with very small sample sizes. For larger models with correspondingly larger sample sizes, ICE is more consistently helpful.

\hfill

For all four configurations, ICE requires less fitting time than MLE, probably due to an earlier exit from the l-bfgs optimizer caused by the less linear nature of the ICE objective function. For the $ [11, 8, 5, 3] $ configuration, fitting time reaches approximate parity for the sample sizes of 32k and above. The computation of the ICE loss and gradient as described here theoretically requires a small constant factor more computation than MLE loss and gradients. This was the experience when performing these tests, though both costs are swamped by other factors and overheads.

\section{Conclusion}

The approach described by \cite{ICE} can be successfully implemented in a MultilayerPerceptron, and should be applicable to any backpropagating neural network using the techniques described here. When implemented in a common off-the-shelf MulitlayerPerceptronClassifier provided by Apache Spark, the result is a classifier that is substantially less susceptible to overfitting for small sample sizes and relatively large parameter counts. The advantage of this approach over $ L_{2} $ and dropout is that it does not require any hyper-parameters, and is therefore viable as a drop-in bias reduction approach.

\pagebreak

\bibliographystyle{abbrv}
\bibliography{mlpIce} 

\end{document}